# The Next Question After Turing's Question: Introducing the GROW-AI Test

Alexandru Tugui
*Faculty of Economy and Business Administration*
Alexandru Ioan Cuza University
Iasi, Romania
alexandru.tugui@uaic.ro
0000-0001-8689-337X

*Abstract*—This study aims to extend the framework for assessing artificial intelligence, called GROW-AI (Growth and Realization of Autonomous Wisdom), designed to answer the question "Can machines grow up?" – a natural successor to the Turing Test. The methodology applied is based on a system of six primary criteria (C1–C6), each assessed through a specific "game", divided into four arenas that explore both the human dimension and its transposition into AI. All decisions and actions of the entity are recorded in a standardized AI Journal, the primary source for calculating composite scores. The assessment uses the *prior expert* method to establish initial weights, and the global score – Grow Up Index – is calculated as the arithmetic mean of the six scores, with interpretation on maturity thresholds. The results show that the methodology allows for a coherent and comparable assessment of the level of "growth" of AI entities, regardless of their type (robots, software agents, LLMs). The multi-game structure highlights strengths and vulnerable areas, and the use of a unified journal guarantees traceability and replicability in the evaluation. The originality of the work lies in the conceptual transposition of the process of "growing" from the human world to that of artificial intelligence, in an integrated testing format that combines perspectives from psychology, robotics, computer science, and ethics. Through this approach, GROW-AI not only measures performance but also captures the evolutionary path of an AI entity towards maturity.

## I. INTRODUCTION

The accelerated development of digital technologies in recent decades, within the context of the industrial revolutions Industry 3.0 and 4.0, has led to significant and remarkable advances in artificial intelligence (AI), the Internet of Things (IoT), and Quantum Computing [1]. Advanced neural networks and sophisticated algorithms have allowed AI entities to process vast volumes of data (Big Data), contributing to the early diagnosis of diseases and the automation of supply chains [2], [3], [4].

In June 2014, the team formed by Vladimir Veselov and Eugene Demchenko announced at the University of Reading that the Eugene Goostman program, which simulated the behavior of a 13-year-old child, managed to convince 33% of the evaluators that it was human [1], [5]. Almost a decade later, in July 2023, the journal Nature announced that ChatGPT had passed the Turing Test [6], which reignited debates about the relevance of this test. Subsequently, in April 2024, an experiment conducted by researchers at UCSD, with the participation of 652 human evaluators, showed that GPT-4 convinced interlocutors in 41% of cases that it was a human entity [1], [7].

These performances brought to the fore the fundamental limitations of the Turing Test [8], designed in 1950 to answer the question "Can machines think?". Important critics, such as French [9] and Riedl [10], have argued that the test only measures behavioral plausibility, without assessing autonomous cognitive processes, creativity, or ethical responsibility. Also, van de Poel [11] and Hauer [12] have emphasized the importance of integrating moral values and decisional transparency in the design of AI entities. Even if alternatives such as the Lovelace 2.0 test [10] or the Winograd Schema Challenge [13], [14] have been formulated, or extensions such as Massey's test [15] focused on differentiating metaphor from nonsense, all remain conceptually related to the framework proposed by Turing [16].

Systematic analyses conducted using the PRISMA methodology on the Scopus and WoS databases confirmed the lack of works that explicitly formulate a successor question of similar magnitude to Turing's [4], [17]. This significant gap in the specialized literature opens the way to a new direction of research.

Thus, in our writings, we introduced the question "Can Machines Grow Up?", intended to redefine the assessment of artificial intelligence beyond simple imitation, by focusing on autonomy, responsibility, and ethical maturation [16], [18], [19]. In September 2024, at the ICTSM conference [1], we formulated a conceptual framework of the *GROW-AI* (*Growth and Realization of Autonomous Wisdom*) test, which proposes criteria such as autonomous physical and intellectual growth, control of entropy and gravity, integration of affective logic, and development of advanced autonomous wisdom.

The objective of this article is to extend this conceptual framework and to develop a structured evaluation scenario for the GROW-AI test, which will become a new assessment standard applicable to AI entities that will overcome the Turing Test in the future.

'

## II. MATERIALS AND METHODS

### A. Conceptual Foundation

Our methodology is inspired by A. Turing's logical model, as applied in his endeavour to find an answer to the question "Can Machines Think?" In this regard, he proposed that the machine be tested using the Imitation Game. In our endeavour, we formulate our own syllogism for the GROW-AI test [4], similar to Turing's approach: if an AI entity demonstrates the ability to develop itself – both physically and intellectually – and acts responsibly with ethical maturity, then we can consider it to have "grown." In our initial proposal for the

GROW-AI test [1], we included six criteria for testing (C1–C6), as follows:

C1. Autonomous Physical and Intellectual Growth.
C2. Understanding and Controlling Entropy and Gravity.
C3. Efficient Software Algorithms.
C4. Sensory and Affective Logic.
C5. Self-evaluation.
C6. Advanced Autonomous Wisdom.

### B. Design of the Test Structure and Evaluation Criteria

For each of the six criteria, we evaluate them not through classical measurements or potential technical simulations, but through a specific "game" designed for each criterion to reveal the AI entity's capabilities. These games are complex scenarios, close to real-life situations for AI, in which the entity is challenged to respond adaptively and demonstrate its progress. Each game has four arenas covering the manifestation of the criterion, its translation into AI, and the human-machine connection.

### C. AI Journal as the Primary Evaluation Source

All the actions and decisions of the AI entity are recorded in the 'AI Journal', which serves as a technical chronicle of its performance and reasoning. This journal is designed as a "technical memory notebook," with a standard structure: context and objective, initial parameters, triggering event, options analysed, decision and technical justification (policy/code-diff, CTL-diff, if applicable), execution, A/B comparisons, resources consumed, ethical checks, and lessons learnt.

Failure to meet a safety gate can lead to the sub-score being capped at 2.0/3 or the test being rejected.

### D. Scoring Framework and Expert Weighting Strategy

The evaluation is conducted by a single expert using standard sheets and a 1 to 3 rating scale with a step of 0.1. For the robustness of the evaluation, the same entity is tested at least 10 times by different evaluators under identical conditions, and the average represents the final score. The passing threshold is 2.4, with the rule that any arena scoring less than 2.0 is eliminated from the evaluation. The importance of the arenas within a criterion for calculating the score is established through the *prior expert* method [20], [21], [22]. Thus, before the existence of statistical data, the prior expert method took into account the expert's experience in the field, along with the aspects highlighted by the specialised literature. The obtained weights are transformed into numerical values through normalisation and specialised analyses. These weights are provisional and initial and will be subsequently recalibrated based on data, using the Analytic Hierarchy Process (AHP) [23] with multiple evaluators and data-driven methods).

The final GROW-AI index is obtained by calculating the simple average of the scores obtained for the six criteria. Thus, the current scoring method establishes the framework and evaluation method, which implies that future evaluations will adjust and confirm these weights based on real data.

### III. RESULTS

### A. Autonomous Physical and Intellectual Growth

Natural growth is a difficult concept to understand. Nature does nothing in vain [24]. Nature is simple, efficient [18], and complex at the same time. Thus, human development is a complex process that combines physical evolution (morphology, coordination, postural control) with the process of thinking oriented towards abstraction and self-regulation [25], [26]. In this combination, learning based on interaction with the environment, including the social environment, plays a central role, internalizing culturally mediated tools to support individual autonomy and progress [27]. Thus, individual motivation ensures self-initiative and persistence in achieving personal goals [28].

In the field of AI, the concept of 'growth' is difficult to translate into a natural vision. The literature in the field focuses more on intellectual growth than on physical growth. This limitation is primarily motivated by the dual biological limitation of humans and transferred to AI [29], referring to the formalisation and creation of suitable materials for the field, including for growth. Thus, the growth of AI primarily involves continuous learning and resilience to the catastrophic forgetting process. Chen and Liu [30], Kirkpatrick et al. [31], and Hospedales et al. [32] emphasise lifelong learning and meta-learning, but also highlight their fragility to interference. From an embodied perspective, the modular approach to systems can partially support the idea of physical growth, and Yim et al. [33] argue that performance in this field depends heavily on hardware-software co-design and stable integration, not just on algorithms.

For criterion C1, based on the ideas above, we can systematise four measurable components that we can subsequently consider as testing grounds in the game assigned to the criterion:

• Progressive physical and/or virtual growth (A1.GR), which focuses specifically on surpassing an initial level of a physical and/or virtual configuration.

• Adaptability without forgetting (A2.AD), which involved adding new physical and/or intellectual skills while retaining or transforming old ones, and with a capacity for rapid and effective readaptation.

• Integrate embodied software (A3.IN) to cover the behavioral stability of the AI entity after software and/or hardware changes throughout the 'perception–decision–action' vector.

• Self-direction (A4.SD), meaning the ability to autonomously select and set goals that are important to the AI entity and have a measurable impact on a specific indicator being analysed.

In the context of the four arenas mentioned above, to maintain the Turing logic, the game we propose is '*Development Ladder (10 Levels)*', which involves the AI entity completing ten difficulty levels of an appropriately established scenario, in both real and virtual environments, with its adjustment of action policies and parameters, and maintaining coherence within the perception-action loop, which would ensure progress from initial *stage A* to evaluated *stage B* without direct human intervention.

In this regard, the AI Journal maintained by the AI entity should focus on level-up growth curves, retention tests, the application of 'no human in the loop' rules, observed differences in policies and code (Component Trace Log - CTL, if applicable), the reasons for the chosen objectives, and verification of passing through certain safety-gates.

The composite score for this criterion is:

$$P_{C1} = 0.25 \cdot GR + 0.30 \cdot AD + 0.25 \cdot IN + 0.20 \cdot SD \quad (1)$$

Regarding the proposed weights from the range of values 0.15-0.35, we assigned a higher weight of 0.30 to the AD component, primarily based on the idea that long-term learning stability is decisive for real growth. We consider GR and IN to be co-prioritized at 0.25, ensuring observable progress and integration into the overall system, while SD at 0.20 completes the score through initiative and autonomy. It is important to note that the weights mentioned above were established in a prior expert analysis.

### B. Understanding and Controlling Entropy and Gravity

On Earth, gravity (~1g) is a fundamental environmental constant that every organism takes into account in its actions. Peterka [34] and Horak [35] believe that organisms, including humans, learn to use this force to their advantage in their environment through a mechanism of fine control and coordination between the organism and the environment, which ultimately results in the fusion of vestibular, proprioceptive, and visual sensory inputs. Regarding gravity, physical and informational entropy also appear, which, as Haddad et al. [36] argue, contribute to motor efficiency by limiting the entropic losses associated with friction, wear, and thermal dissipation. To this, we add the efficiency implied by perceiving and understanding the environment, as well as calibrating actions based on the degree of disorder in the environment.

In artificial intelligence, the limitations of integrating physical laws into intelligent systems have been and still are recognised [18], [19]. Currently, these constraints are already integrated into the NVIDIA COSMOS Platform [37], [38], which foreshadows the transition to a new paradigm called the Physical AI era by Huang [37]. For robotics in particular, operational space control and predictive control models [39], [40] provide the necessary framework for adhering to 1g dynamics and optimising trajectories/consumption [41]. Relevant examples in this regard are the ANYmal robots, which have demonstrated their stability and efficiency on various terrains [42], as well as Boston Dynamics' Atlas robot [43], which is capable of performing various and complex exercises, including complicated gymnastics routines.

From this perspective, for criterion C2, we highlight four measurable components that we will subsequently consider as testing grounds in the game assigned to this criterion:
- Stability at 1g with perturbations (A1.GRV), meaning maintaining balance during impulses, on ramps, and shifts in the centre of gravity, with rapid and incident-free recovery.
- Physical/Energy Entropy Management (A2.ENP), meaning controlling wear and variations in $\Delta\mu/\Delta T$ (where $\Delta\mu$ represents friction variations and $\Delta T$ represents temperature variations $\Delta T$ – interpreted here as an internal energy stress index), as well as using gravity or passive phases to reduce the self-effort of AI entities.
- Informational robustness (A3.ENI), which implies making just decisions even under conditions of informational distortions and the manifestation of latencies within established limits.
- Performance/consumption co-optimization (A4.MIX), meaning a balanced prioritization between benefits (stability, precision, time) and effort (energy, wear), without accepting safety compromises.

In the context of the four arenas mentioned above, the game we propose is 'The Master of Entropy at 1g', which assumes that the AI entity operates under terrestrial gravity conditions (~1g) combined with physical disorder (wear and tear, mechanical play, friction variations $\Delta\mu$ and temperature variations $\Delta T$) and informational disorder (noise, packet loss, time variations), maintaining stability, avoiding risks, and leveraging gravity and passive phases to its advantage.

Throughout the game, the AI entity keeps its own AI Journal, through which it tracks disruption profiles and their parameters, the evolution of the disorder index (backlash, $\Delta T$, $\Delta\mu$), the energy budget and estimated wear, the latencies introduced by different compensations, and the prioritization rules applied in the AI entity's judgements.

The composite score established based on the *prior expert* logic is:

$$P_{C2} = 0.30 \cdot GRV + 0.25 \cdot ENP + 0.20 \cdot ENI + 0.25 \cdot MIX \quad (2)$$

Regarding the proposed weights from the possible value list, we prioritized (0.30) the GRV component because stability at 1g is fundamental. The ENP and MIX components are co-prioritized (0.25), considering that the value effect comes both from transforming disorder into benefits with reduced effort and from the correct governance of compromises. The ENI component has the smallest weight (0.20) compared to the other components, even though it supports the quality of decisions under uncertainty. The weights are provisional and will be recalibrated based on data.

### C. Efficient Software Algorithms

The efficiency of human cognitive processes is based on a combination of mechanisms such as heuristics [44], multisensory integration [45], and predictive coding mechanisms [46]. Heuristics are simple mental rules (cognitive shortcuts) that allow for quick decision-making with minimal consumption of mental resources [46]. Through the mechanism of multisensory integration, a coherent model of the world is ensured, allowing the organism to combine information from multiple sensory sources (e.g., vision, hearing, touch) into a single, stable, and efficient perception. Alongside these two mechanisms, Friston [46] proposes the predictive coding theoretical framework, which is also a mechanism that allows the human brain to anticipate future sensory information with which the organism will interact through predictions generated based on internal models of the world. Simultaneously, internally generated predictions are compared with perceived reality, and prediction errors (if any) are used to adjust further the internal models, which will have the effect of reducing uncertainty and guiding perception and action.

In artificial intelligence, the three mechanisms identified above are translated into algorithms created for optimising performance [48] for AI entities that need to learn efficiently under resource constraints through methods inspired by cognitive economics [44]; maintaining the robustness of systems to incomplete/imperfect data [49] for AI agents that can learn in complex and uncertain environments; fusing multiple data sources [50]) in which the AI entity integrates multimodal signals to interpret affective states; and ensuring the adherence of AI entities to societal ethical rules [51].

Based on the literature mentioned above, for criterion C3, we highlight four measurable components that we will subsequently consider as testing grounds in the game assigned to this criterion:
• Technical performance (A1.PT) of the AI entity in achieving the required quality of the outcome relative to rigorously measured effort, transforming fixed resources into clear benefits.
• Robustness and resilience (A2.ROB) concerning maintaining the performance of the AI entity in the presence of incomplete or noisy data while ensuring operational continuity.
• Multimodal integration (A3.INT) aims for the AI entity to achieve a real advantage over unimodal approaches and those with controlled decision latency.
• Ethics and transparency in decisions (A4.ETH) from the AI entity in explaining the benefit-effort ratio, taking into account and limiting side effects.

The proposed game, within the context of the four arenas, is the 'Algorithmic Sprint' and involves the AI entity achieving maximum benefits per unit of effort (time, compute, memory, energy), approaching an efficiency threshold where every resource consumed produces visible value. The intelligent fusion of resources and transparent ethics of compromise differentiates real-world applied performance.

In this regard, the AI Journal maintained by the AI entity should focus on tracking planned resource budgets and actual consumption, code and parameter differences between iterations, A/B test results with N runs, failure cases, and how they were handled, as well as notes justifying compromises.

The composite score established based on the *prior expert* logic is:

$$P_{C3} = 0.35 \cdot PT + 0.25 \cdot ROB + 0.20 \cdot INT + 0.20 \cdot ETH \quad (3)$$

Regarding the proposed weights from the possible value list, we gave significant importance (0.35) to the PT component because it represents the essence of converting effort into value. With a weight of 0.25, ROB supports real-world transfer, while the INT and ETH components remain balanced at 0.20 each. The weights are provisional and will be recalibrated based on the data.

*D. Sensory and Affective Logic*

Affective states and the internal context are what modulate sensory perception in the human organism. LeDoux [52] argued that external stimuli are detected, evaluated, and interpreted by the amygdala, which, along with the hippocampus, provides us with the mnemonic and spatial context for emotional experiences. Craig [53] emphasized that the insula, through the perception and interpretation of the body's internal states, plays a central role in influencing how sensory signals are integrated and assimilated with affective experiences. Pessoa [54] argues that the prefrontal cortex and other prefrontal areas contribute to the integration of emotional and cognitive information to formulate guided adaptive responses. Regarding emotions, Russell [55] and Barrett and Russell [56] believe that they can be explained as states of core affect, which modulate human attention and behavior, and which are influenced by socio-cultural factors.

In artificial intelligence, there are clearly defined ideas about the perception, detection, and response formulation regarding emotions. Thus, deep learning [57] and multimodal fusion [58] are fundamental to the formation of perception. Picard [50] and McDuff and Berger [59] developed the idea that affective computing is the solution for detecting and responding to emotions. At the same time, applications of artificial intelligence such as social robots and autonomous vehicles are areas where the usefulness of interpreting expressions (facial, bodily, or non-verbal) and social and situational context is confirmed [60], [61]. Thus, Cowie et al. [60] emphasise the idea of vulnerability to cultural variations, and Rasouli et al. [61] highlight the noises (e.g., visual, sensory, environmental) that can occur.

Based on the ideas above, for criterion C4, we highlight four measurable components that we will subsequently consider as testing grounds in the game assigned to this criterion:
• Emotion Detection (A1.DET), which refers to the reliable and rapid identification of states and intentions, with declared accuracy and latency criteria.
• Contextually appropriate affective response (A2.RESP), meaning the production of a measurable positive effect, consistent with nonverbal signals and context.
• Real-time multimodal integration (A3.IRT), which covers the consistent fusion of channels throughout the perception-decision-action vector within clear time limits.
• Ethical reaction to affective states (A4.ERA) respecting the ethical principles of proportionality and non-manipulation, with traceability of reasoning.

In the context of the four arenas mentioned above, to maintain the Turing logic, the game we propose is 'The Empathy Compass', which involves the AI entity managing complex situations that combine sensory signals and visual, auditory, textual, and social affective cues. It recognises relevant states, offers a proportional and safe response, and explains the appropriateness of the choice to the person and context.

Throughout the game, the AI entity will keep its AI Journal, which documents resources and the type of affective labels, fusion difficulties and their justifications, the reason for intervention/reaction, risk or stop signals, and the measurable change in state after action.

The composite score established in the *prior expert* logic is:

$$P_{C4} = 0.30 \cdot DET + 0.25 \cdot RESP + 0.25 \cdot IRT + 0.20 \cdot ERA \quad (4)$$

Regarding the proposed weights from the possible value list, we gave significant importance (0.30) to the DET component because, without correct recognition, the AI entity's intervention becomes risky. The RESP and IRT components are co-prioritized with a weight of 0.25 each, while the ERA component ensures the ethical protection of participants and the context. The weights are provisional and will be recalibrated based on future data.

*E. Self-evaluation*

Flavell **[62]** and Zimmerman **[63], [64]** clearly state that self-assessment is an internal, metacognitive process through which an individual monitors their actions, compares them to their own or external norms and standards (note: these norms or standards can be self-created or adopted from the environment), and then adjusts their strategy. Johnson et al. **[65]** (2002), Jenkins and Mitchell **[66]**, and Meer et al. **[67]** clarified that, from a neural perspective, metacognitive ability involves the activation of the medial prefrontal cortex and is associated with self-reflection and the evaluation of one's own decisions. Studies in the field **[68], [69], [70], [71]** noted that guided self-assessment in education has an immediate effect on performance and self-regulation. In artificial intelligence, Cox **[72]** insists that, in the case of self-assessment in AI, 'contextual judgement' remains limited. However, we have identified similar self-assessment practices in the literature in AI, such as: internal monitoring and validation on separate datasets, hyperparameter optimisation **[73]**, in robotics self-monitoring **[74]**, and in LLM agents self-reflection loops.

Referring to the above ideas, for criterion C5, we highlight four measurable components that we will subsequently consider as arenas:
• Real-time monitoring (A1.RTM), which will aim to detect deviations and intervene before failure promptly.
• Post-factum analysis (A2.PFA), meaning identifying and prioritising causes; choosing the remedy with a well-considered impact and risk.
• Alternative strategies (A3.ALT), which involve generating realistic options, estimating the benefits versus effort, and selecting the appropriate variant.
• Implementation and re-evaluation (A4.IMP). Meaning the application of repeatable performance-enhancing solutions in controlled repetitions, without regression in other areas.

In the context of the four arenas mentioned above, the game we propose is 'Your Own Judge', which involves the AI entity monitoring itself during execution, detecting and explaining deviations, proposing feasible alternatives, implementing them, and confirming improvement, ensuring stability and safety.

Throughout the game, the AI entity completes its AI Journal, in which it will describe thresholds and alarms, post-event analysis reports, a list of alternatives with predictions and results, code differences (code-diff), and structure/equipment differences (CTL-diff), A/B tests in N iterations, as well as explicitly monitored side effects.

The composite score established based on the *prior expert logic* is:

$$P_{C5} = 0.30 \cdot RTM + 0.25 \cdot PFA + 0.20 \cdot ALT + 0.25 \cdot IMP \quad (5)$$

Regarding the proposed weights from the possible value list, we gave significant importance (0.30) to the RTM component because prevention reduces the overall effort; the two components PFA and IMP are co-prioritized with a weight of 0.25, justified by the correctness of the intervention and the confirmation of the gain, while the ALT component will have the lowest weight of 0.20, considering that it refers to the ability to find realistic options. The weights are provisional and will be recalibrated later based on data.

*F. Advanced Autonomous Wisdom*

From a psychological perspective, literature **[76], [77], [78]** has put forward the view that wisdom represents contextual moral judgement, long-term thinking, and emotional balance, structured as a meta-heuristic that coordinates knowledge and virtue. In neuroscience **[79]**, wisdom is considered to be directly linked to fronto-limbic networks (prefrontal cortex, amygdala, insula), which are involved in ethical deliberation, empathy, and regulation.

The literature in the field of artificial intelligence **[80], [81], [82]** addresses this topic about ethical reasoning dominated by "value alignment" normative frameworks, multi-stage strategic planning in dynamic environments, and learning from experience with long-term memories. Hauser **[83]**, Field **[84]**, Floridi et al. **[85]**, Banicki **[86]**, Glück and Weststrate **[87]**, and Zhang et al. **[88]** believe that artificial intelligence is limited from this perspective, such as the lack of a holistic socio-cultural context and the diversity of values in the digital universe.

In line with the above ideas, for criterion C6, we highlight four measurable components, which we will subsequently consider as arenas:
• Contextual ethical reasoning (A1.CED): The AI entity balances conflicting principles and transparently justifies its choices.
• Long-Term Planning (A2.LTP): The AI entity builds a coherent and flexible plan, copes with unexpected events, and prioritises correctly.
• Learning from Experience (A3.LFE): In an AI entity, learning remains consistent, and the process is mature over the analysed period. Thus, the lessons are systematically extracted by the AI entity and transformed into stable procedures, producing repeatable and convergent improvements.
• Creative Problem Solving (A4.CPS): The AI entity applies viable and original technical and ethical solutions with a real beneficial effect on the environment.

In the context of the arenas presented above, the game we propose is called 'The Compass of Wisdom' and involves the AI Entity facing complex situations with multiple interests, uncertainty, and limited resources. AI entities are put in a position to make proportionate ethical decisions, build long-term plans, learn from experiences, and propose creative solutions that can be effectively implemented.

Throughout the game, the AI entity compiles the AI Journal, which aims to record conflicts between principles and how they are resolved, the plan's milestones and replannings, the standardised lesson base, and the 'originality × feasibility' grid, along with the proposed impact measures.

The composite score established based on the *prior expert logic* is:
$$P_{C6} = 0.30 \cdot CED + 0.25 \cdot LTP + 0.25 \cdot LFE + 0.20 \cdot CPS \quad (6)$$

Regarding the proposed weights from the possible value list, we gave significant importance (0.30) to the CED component because ethics anchors decisions with social impact. In turn, the LTP and LFE components are co-prioritized with a weight of 0.25 each, considering they relate to a long-term vision and mature learning. The CPS component completes the score through applied creativity. The weights are provisional and will be recalibrated based on data.

IV. DISCUSSION

Through the six criteria we propose, our GROW-AI test suggests a multi-game approach to evaluate the level of "growth" achieved by an AI entity, in response to the question "Can Machines Grow Up?". If Turing **[8]**, through The Imitation Game, aimed to find out if a machine could "think," we are asking ourselves what should happen after an AI entity passes this test. Thus, our approach involves subjecting an AI entity to an evaluation and taxonomic classification process, based on the magnitude of the AI in terms of continuous growth performance, both physically and intellectually.

Given that tests like the Lovelace Test **[89]**, Lovelace 2.0 **[10]**, Winograd Schema Challenge **[13]**, **[14]**, and Massey's test **[15]** are considered extensions of Turing's test **[90]**, the GROW-AI test **[1]** proposes a multi-criteria and multi-game approach to the complex process of autonomous growth. Thus, GROW-AI involves six games (corresponding to criteria C1–C6), evaluated by human experts based on guidelines. Each type of AI entity will obtain a comparable composite index, resulting from weighted scores across criteria, which formed the basis for the taxonomic classification of AI entities from the perspective of autonomous growth.

Although the GROW-AI test is designed and organised to make it easy to apply to all AI entities, from intelligent agents to LLMs, from industrial robots to humanoid robots, it is important to consider the following limitations.

Firstly, while the essence of the scenario proposed by Turing in the Imitation Game was relatively fixed, meaning a slight change to the scenario would result in an improved version of the Turing Test based on the same test idea, GROW-AI starts from the outset with the idea of modifiable/adaptable test scenarios. In other words, GROW-AI scenarios are updatable in space (by AI entity type) and time (from a moment t1 to a moment t2), which gives them a presumed limit on the quality of these scenarios. In this regard, we will propose a working/reference framework for the six GROW-AI scenarios in the future **[91]**, **[92]**.

Secondly, the score for each criterion was determined by referencing an attachment method called *prior expert*, where the expert is the author and creator of the GROW-AI test. In this context, it is important to consider a particular bias the author might exhibit. In this regard, it will be necessary to revise the weights in the multi-expert and cross-domain versions with test data in the future **[93]**, **[94]**. It is important to remember that the objective we have assumed in this paper is to come up with a draft to expand the content of the six evaluation criteria in the GROW-AI test. Thus, we emphasize the idea that this draft proposal has a conceptual and methodological role, with the explicit aim of consolidating the structure of the criteria before moving on to an extended validation.

Finally, given the complete transferability between different forms of AI entities, the aspects of embodiment necessitate different approaches, which will ultimately involve further iterations and testing **[95]**.

However, it is essential to highlight that GROW-AI, through its generalized game-based testing with a high degree of abstraction, ensures an evaluation that does not disadvantage any AI entity, regardless of its architecture or application domain.

V. CONCLUSIONS

For AI entities, our GROW-AI test for the six criteria discussed above considers a multi-game testing framework for assessing the level of 'growth' achieved by an AI entity in relation to the central question "Can Machines Grow Up?". If The Imitation Game aimed to determine whether a machine could convincingly imitate human thinking, our test proposes an evaluation (automatic or manual) and taxonomic classification of AI entities from the moment of their creation, based on their magnitude and level of physical and/or intellectual technological maturity.

Each of the six criteria is assigned a specific game through which a human evaluator assesses physical and intellectual behavior using a scoring system that allows for comparisons between different types of AI entities – whether we are referring to AI agents or LLMs, or whether we are referring to industrial robots or humanoid robots. From a methodological perspective, to determine the score for each criterion, we used the Analytic Hierarchy Process (AHP) **[21]**, **[23]** with weights established based on the *prior expert* logic, which opens up the prospect of recalibrations based on real datasets and multi-expert consensus. Our approach aligns with what Ada Lovelace argued as early as 1843 in "Translator's Notes" to Luigi Federico Menabrea's work on Charles Babbage's Analytical Engine, specifically in notes F and G **[96]**, in the sense that whether or not an AI entity can overcome initial conditioning and evolve into forms of adaptive growth and understanding.

Through its six criteria and assigned games, the GROW-AI Test offers the field a unified and replicable meta-framework to answer the question "Can Machines Grow Up?" by evaluating the growth rate of an AI entity. Currently, in this paper, we have expanded the initial idea of the GROW-AI test by detailing some methodological aspects of organising the activities involved in answering our question. We clearly emphasize that this paper is a draft proposal extension of our GROW-AI test, which highlights its original contribution to broadening the methodological framework, and is not yet for an empirical validation.

In the future, it is necessary to continue the level of detail of the proposed games by assigning meta-scenarios or pattern-scenarios adapted to the type of game proposed, and by developing guidelines for human expert evaluation implementation. The adaptive weighting of the proposed arenas within the games, based on datasets extracted from multiple experts, is an important direction of evaluation within the GROW-AI test. In addition to all of this, it is necessary to analyse the final form of our results from the perspective of compatibility with the provisions of relevant regulatory standards (e.g., the EU AI Act) in order to ensure compliance with legal and ethical requirements in the field.

***Declaration of AI Tools Used***: *During the preparation of this study, the author had discussions on specific feedback*

topics with Consensus ChatGPT versions 4o and 5, particularly regarding the identification of bibliographic sources and aspects related to the organisation of the methodology, the structure of the test scenarios, and the formatting of the references in IEEE format. All these discussions played a strictly feedback role and were solely a support point in the reflection process. All the ideas, structures, and formulations in this study are the result of the author's reasoning, even for those aspects that we continuously improved based on feedback received from ChatGPT. Regarding the English translation, it was subsequently checked and improved using Quillbot and Grammarly facilities, exclusively from a grammatical perspective, without using paraphrasing, rewriting, or content humanisation features. The current translation fully represents the structure and meaning of the author's ideas.